# Conveying Imagistic Thinking in TCM Translation: A Prompt Engineering and LLM-Based Evaluation Framework

Jiatong Han[a]*

[a]*Department of Philosophy, Zhejiang University, Hangzhou, China*

Email: joanwahson@gmail.com

Abstract: Traditional Chinese Medicine (TCM) theory is built on imagistic thinking, in which medical principles and diagnostic–therapeutic logic are structured through metaphor and metonymy. However, existing English translations largely rely on literal rendering, making it difficult for target-language readers to reconstruct the underlying conceptual networks and apply them in clinical practice. This study adopted a human-in-the-loop (HITL) framework and selected four passages from the medical canon Huangdi Neijing (黄帝内经) that are fundamental in theory. Through prompt-based cognitive scaffolding, DeepSeek V3.1 was guided to identify metaphor and metonymy in the source text and convey the theory in translation. In the evaluation stage, ChatGPT 5 Pro and Gemini 2.5 Pro were instructed by prompts to simulate three types of real-world readers. Human translations, baseline model translations, and prompt-adjusted translations were scored by the simulated readers across five cognitive dimensions, followed by structured interviews and Interpretative Phenomenological Analysis (IPA). Results show that the prompt-adjusted LLM translations perform best across all five dimensions, with high cross-model and cross-role consistency. The interview themes reveal differences between human and machine translation, effective strategies for metaphor–metonymy transfer, and readers' cognitive preferences. This study provides a cognitive, efficient and replicable HITL methodological pathway for translation of ancient, concept-dense texts like TCM.

## Introduction

The translation of Traditional Chinese Medicine (TCM) classics concerns a body of historical texts built upon highly complex theoretical systems. A defining feature of ancient medical thinking is its reliance on imagistic cognition, where dense metaphorical and metonymic expressions function as cognitive cues for clinical reasoning (Jia *et al.*, 2014; Jiang *et al.*, 2022; Han *et al.*, 2024). The culturally embedded and largely implicit logic of these texts, however, is difficult to discern and articulate, making it challenging for target-language readers to fully comprehend and apply TCM principles in clinical practice.

As classical medical texts seldom make their metaphorical and metonymic mappings explicit—and instead rely on them as a cognitive background—translators must identify these mappings and restore their narrative functions in translation. Only then can target-language readers grasp the underlying medical reasoning and experience the "imagistic thinking" that structures TCM. The translation therefore needs to move beyond semantic fidelity. It must attend to the metaphorical grounding of medical concepts (Lakoff & Johnson, 2011) and the metonymic patterns that organise relations among events and phenomena (Kövecses & Radden, 1998). In other words, the task is not simply linguistic but cognitive: not merely the rendering of terms, but the reconstruction of the conceptual schemas through which ancient physicians understood the body and its processes (Johnson, 1987; 2007).

Most existing translations rely on literal rendering. Although concise and fluent, such translations often compromise clarity (Chen *et al.*, 2025). Classical Chinese operates without explicit connective markers, and its clinical logic depends heavily on metaphor–metonymy relations that function as cognitive linkages rather than surface linguistic structures. Target-language readers, however, generally lack the conceptual foundations embedded in ancient medical practices and philosophical systems (*e.g.*, the Five Phases, yunqi cosmology). Without sufficient explanation, literal translations obscure the reasoning structure of the text. By contrast, LLM-generated translations tend to be more explanatory and reader-oriented, thereby reducing cognitive load. Yet they still fall short in detecting and reconstructing metaphor and metonymy, making them inadequate for supporting deeper cognitive needs—understanding, transferring, and applying TCM principles.

Existing research in translation evaluation has largely centered on terminological accuracy, comparative analyses of prior translation strategies, and the performance of LLMs（Wang & Chen, 2023), whereas far less attention has been directed toward readers' cognitive pathways and cognitive needs—an area that remains insufficiently explored but is crucial for understanding how translations function in practice. As TCM classics are fundamentally medical texts rather than cultural artifacts, their primary purpose is to guide clinical reasoning and inform therapeutic decision-making. The dissemination of medical principles therefore requires not only terminological precision but also translation that enables readers to reconstruct diagnostic logic and to develop confidence in applying it in clinical settings. Accordingly, translation quality should be assessed not merely in linguistic terms, but also in terms of its pragmatic efficiency and cognitive depth (House, 2015; Xiao & Muñoz Martín, 2021), particularly regarding how well it assists readers in navigating and internalizing the conceptual structure of TCM.

Against this background, this study examines Huangdi Neijing—the most frequently translated TCM classic—and proposes a human–machine collaborative approach to improving the translation of metaphor- and metonymy-laden passages. Using prompts as cognitive scaffolding, TCM experts supply the background knowledge needed to steer the LLM toward identifying relevant metaphor–metonymy cues and adjusting its renderings accordingly. The LLM, in turn, internalizes the prompted reasoning patterns and incorporates them into its translation output. Four passages from the Huangdi Neijing—dense in metaphor and metonymy, difficult to interpret, and central to theoretical guidance—were selected for this translation experiment. For evaluation, a mixed-method design was adopted, in which LLMs functioned as computationally simulated readers that compared human, baseline, and prompt-adjusted translations from pragmatic and cognitive perspectives, providing explicit reasoning to reveal their reception pathways. The design is shown in Figure 1.

Figure 1. Flowchart of the study design.

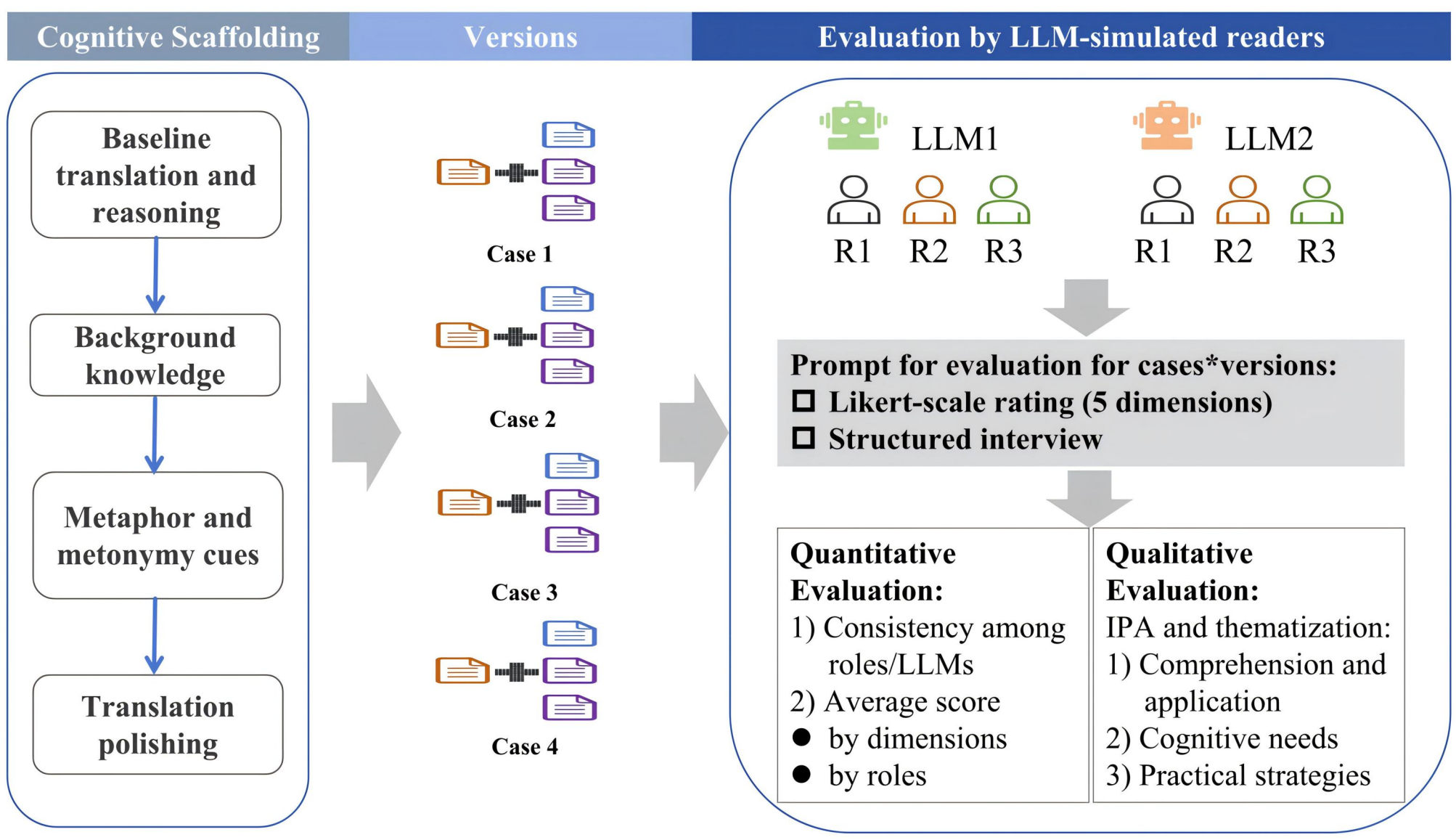

(Note: Each case has four versions. Red: prompt-adjusted translation; blue: baseline translation; purple: two human translations.)

This study addresses the following research questions: (1) How can translations enhance cognitive transferability, enabling readers to accurately understand and apply TCM principles? (2) What are the cognitive needs of different types of readers, and where do they encounter obstacles in comprehending and applying existing translations?

(3) Which translation strategies for TCM classics enhance cognitive efficiency, and how can LLMs be leveraged to support these strategies?

The contribution of this study is treating LLMs not merely as translation tools but as adaptive systems capable of rhetoric learning（Shao *et al.*, 2024; Tang *et al.*, 2025）. Through prompt engineering, the cognitive patterns embedded in TCM are transferred to the model, resulting in comprehensive improvements in translation quality. The findings highlight the explanatory power of metaphor and metonymy as cognitive cues in TCM reasoning and their critical role in facilitating cognitive transfer during translation. By providing a reproducible and efficient framework, this study advances the cross-cultural communication of TCM theory, extends theoretical understanding of AI-assisted translation within translation studies, and offers guidance for translating historically and conceptually complex texts. The following sections review relevant literature, research design, quantitative and qualitative results, and the theoretical and practical implications of this approach.

## Literature review

### *2.1 Translation of TCM Classics and Its Challenges*

Classical TCM texts contain complex conceptual systems and characteristic modes of thought. Their translation is highly valuable yet extremely challenging, functioning as a representative case of texts that are simultaneously theoretical, historical, and cultural. Since the 1990s, translation efforts for TCM theories have gradually expanded, largely focusing on the standardization of terminology and the translation of classical texts, with English translations dominating the field.

Terminology studies have progressed more rapidly than classical texts for the growing clinical needs (Lim *et al.*, 2022; Feng *et al.*, 2024). Currently, the two best known versions are the WHO International Standard Terminologies on Traditional Chinese Medicine (WHO, 2022) and the International Standard Chinese-English Basic Nomenclature of Chinese Medicine issued by the World Federation of Chinese Medicine Societies (Li, 2008). The standards formed the basis of clinical guidelines, diagnostic protocols, and international medical collaboration.

In contrast, the translation of classical TCM texts is more individualized and influenced by the translator's expertise and style. Since high-quality translations of canonical texts are few, the scale of translation research is limited, and the translation theories adopted are scattered and lack a unified paradigm. Among classical texts, Huangdi Neijing has received most attention, the two most academically influential translations are those by Li Zhaoguo and Paul U. Unschuld. Li, as a native-language translator, prioritizes fidelity to the linguistic and stylistic characteristics of the original, whereas Unschuld—working from German, a West Germanic language closely aligned with English—can be regarded as a target-language–oriented translator whose renderings display poetic expressiveness and a natural idiomatic flow.

Because of its cultural complexity, the translation of TCM texts faces a dilemma between overt and covert strategies. For example, Lim *et al.* argue that while literal translation preserves terminological integrity and free translation enhances clarity, the two aims can hardly coexist, so a context-dependent strategy might help (Lim *et al.*, 2022). Chen *et al.* report, through student evaluations, that domestication in translation leads to easier but compromised understanding of terms (Chen *et al.*, 2025). Translators of Huangdi Neijing note that, due to the cultural and historical gap between the source and target text, the greatest challenge is conveying cultural concepts and transferring cognitive modes. Therefore, a shared strategy is to supplement the text with extensive commentaries from later physicians, which assumes that it is the reader's responsibility to learn, discern and differentiate the theoretical information in the annotations (Li, 2008; Unschuld, 2016). This is unrealistic for target readers who lack TCM background and historical knowledge, and likely to make the reading burdened and confusing.

What is worth noting is that the above studies demonstrate a reader-oriented perspective, examining whether the translation helps readers understand and apply TCM theory. This represents a valuable pragmatic turn given the clinical importance of TCM treatises. As House emphasizes in her pragmatics-oriented translation evaluation framework, translation constitutes a cognitively grounded transfer that occurs both in the translator's mind and in cross-linguistic and cross-cultural social practice: "Translation is both a cognitive procedure which occurs in a human being's, the translator's, head, and a social, cross-linguistic and cross-cultural practice. Any valid theory of translation must embrace these two aspects." (House, 2015, p. 1). Thus, the core task of TCM translation is to convey the underlying cognitive patterns of TCM.

### *2.2 Cognitive Features of TCM Language*

Since around 2010, researchers have noted the compatibility between cognitive linguistics and TCM thought (Jia, 2012). The cognitive model of TCM is derived from sensory and practical experience and is highly imagistic, reflected linguistically through extensive use of metaphor and metonymy—metaphor for explanation and metonymy for connection. Semantics is essentially the expression of imagistic thinking, in which image schemas grounded in embodied experience are constructed and abstracted (Shi & Yue, 2024; Liu *et al.*, 2025), serving as source domains for metaphors that interpret and categorize medical concepts (Lakoff & Johnson, 2011). These schemas are drawn not only from basic spatial and temporal perception (Zhong & Liu, 2022) but also from meteorology, agriculture, astronomy, geography, warfare, and governance, explaining physiological and pathological mechanisms in a dynamic, philosophical, and holistic way (Jia, 2014).

Once metaphors become conventionalized as theoretical categories, metonymy becomes an additional powerful tool. Metonymic relations are organized around the essential features of categories and events, and they build cross-ontological and cross-domain connections according to context (Kövecses & Radden, 1998). TCM learners first acquire a concrete understanding of medical categories through metaphor; later, in clinical discourse and practice, they employ metonymic reference to connect categories and events, forming a coherent theoretical understanding that guides clinical reasoning (Han *et al.*, 2024).

These cognitive structures become transferable once metaphor and metonymy—as their linguistic representations—are identified and interpreted. This is supported by cognitive linguistics and cognitive psychology, which emphasize that humans form image schemas and idealized cognitive models as reference structures for organizing experience and adapting to the environment, corresponding to neural connectivity patterns (Johnson, 1987; Kövecses & Radden, 1998; Lakoff & Johnson, 2011). As these cognitive structures are highly flexible and well adapted to the environment, they can be understood across cultures through shared embodied experience (Johnson, 2007; Zhou & Wang, 2025). In effective translation, such metaphor adaptation and semantic frame shifting help overcome cultural and idiomatic barriers (Ali & Jameel, 2025). Therefore, the cross-cultural challenge of TCM translation is to convey the schemas rather than the surface text. Recent studies highlight the role of metaphor and metonymy in enabling embodied cognition transfer and cultural re-enactment in translation (Zhou & Wang, 2025; Jiang *et al.*, 2022). However, research on metonymy remains far less developed than research on metaphor. Given the historical and cultural distance, conveying metaphors is difficult in practice, and their translatability is limited (Hong & Rossi, 2021; Kuang & Tang, 2025). The solution is to incorporate metonymy: once metaphors "die" and become conceptual systems—such as the Five Phases framework in TCM—their imagistic features evolve into the basis for referential relations in metonymy (Kövecses & Radden, 1998), which constitutes the logic used in clinical practice. Accordingly, this study employs both metaphor and metonymy as fundamental elements for adjusting LLM translations to optimize them at the cognitive level.

### *2.3 Optimization of LLM Translation*

LLMs function primarily as tools or assistants in translation. Trained on massive corpora, they possess strong linguistic capabilities, produce translations, and enhance translators' proficiency in the target language. However, their cultural sensitivity, professional knowledge, and contextual awareness are still limited compared with those of human translators (Shahmerdanova, 2025), and translation quality is affected by disparities in target-language resource availability. Jabak identifies two primary areas where LLMs support human translators: AI-Augmented Decision Making (AADM) and Ethical & Bias Mitigation Layer (EBML), with high-quality output depending on human feedback grounded in context, culture, and domain expertise (Adaptive Feedback Loop, AFL) (Jabak, 2025; Jiao *et al.*, 2023).

The optimization of LLM translation usually proceeds in the following aspects. Low-resource and exploratory studies typically rely on prompt-based methods to probe model capabilities and transferability, whereas mature and well-structured datasets employ fine-tuning for stable performance improvement.

Domain adaptation and application. This is common in medical translation, such as zero-shot prompt-based tuning to improve readability in health communication (Githinji *et al.*, 2025) or enhance terminological accuracy (Rios, 2025).

Cultural and contextual supplementation. Supplementing background knowledge is essential in this context. For example, Ranade *et al.* demonstrate that prompts encoding rhetorical and contextual information improve model comprehension and user interaction, with applicability to pedagogical contexts (Ranade *et al.*, 2025). N-shot prompt injection allows models to quickly learn contextual roles and discourse styles, producing human-like translations (Pourkamali & Sharifi, 2025).

Metaphor learning and cognitive scaffolding. LLMs lack robust capability to recognize metaphor, especially when the role of metaphor is not rhetorical but conceptual (Guo, 2023). For this, Shao *et al.* incorporated metaphorical interpretations into fine-tuning datasets as chain-of-thought (CoT) modules, which significantly enhanced the translation quality (Shao *et al.*, 2024). In TCM, Tang *et al.* analyzed metaphors in perceptual dimensions (familiarity, emotional arousal, semantic accuracy) , which served as an effective reasoning scaffold for LLMs—improving DeepSeek V3 accuracy by over 20%. They suggest domain-tailored parsing to enhance domain adaptation (Tang *et al.*, 2025).

As seen in the above approaches, human-in-the-loop (HITL) is an effective methodology in enhancing LLM translation, which could either be AI-led with human supervision (HITL) or human-led with AI assisting with repetitive tasks (AITL) (Natarajan *et al.*, 2025). Both modes consider humans and LLMs as co-translators: the former for cultural and domain understanding, the latter for readable expressions and large volumes of work (Lianeri & Zajko, 2008; Vassallo, 2015; Latif & Arshad, 2025). This suggests LLMs and humans are both interpretive mediators, which takes the knowledge and feedback from human expertise as high-quality contextual databases for low-resource texts (Yang *et al.*, 2023). Such reader-oriented translation corresponds with cognitive translation theory that is integrating with psychological methods (Xiao & Muñoz Martín, 2021).

### *2.4 Evaluation of LLM Translation*

As LLM-based translation and its optimization increasingly focus on cognitive transferability and contextual interpretability, evaluation frameworks have likewise become more effective and practical. Current studies combine automatic and human-based approaches and introduce human-like, novel automatic evaluation models.

In existing research, automatic evaluation metrics rely on reference translations and remain less sensitive to context and culture. BLEU (Papineni *et al.*, 2001), based on lexical similarity, offers limited value. Neural evaluation models assess semantic consistency more robustly (Freitag *et al.*, 2022). For example, COMET and BERTScore measure semantic similarity (Rei *et al.*, 2020; Zhang *et al.*, 2020), and BartScore evaluates generated text quality (Yuan *et al.*, 2021). All of these require high-quality reference translations (Qian *et al.*, 2024). Meanwhile, human evaluation remains largely experiential, focusing primarily on comparing translation strategies in existing renditions (Lim *et al.*, 2022).

These metrics are insufficient for TCM translations. TCM classics are extremely low-resource and zero-reference texts, and their translations are cognitively open, requiring flexible evaluation. For such low-resource scenarios without reference translations, models such as GEMBA employ multi-round back-translation to compare source–target similarity, reducing dependence on references (Kocmi & Federmann, 2023). Document-level and long-context evaluation methods have also been shown to approximate human assessment more closely (Haq *et al.*, 2025). For culturally or contextually sensitive assessments, expert contextual intervention remains essential (Huang *et al.*, 2024). Qualitative methods and context-sensitive frameworks such as House's model (House, 2015) highlight differences in cultural interpretation and cognitive load, offering deep insights into the explanatory function of translation (Chen *et al.*, 2023; Metwally *et al.*, 2025; Mohammed, 2025) and guiding the formulation of context-appropriate strategies (Calvo & De La Cova, 2023).

Although more thorough and instructive, qualitative and context-sensitive evaluation typically involves small samples and limited generalizability. However, LLM-as-a-judge, as a novel evaluation design, may overcome these limitations (Yuksel *et al.*, 2025; Baysan *et al.*, 2025). With context-adjusted prompts, LLMs can act as evaluators to rate comprehensibility, consistency, and pragmatic adequacy, though their performance remains sensitive to prompt design and model type (Qian *et al.*, 2024; Bavaresco *et al.*, 2025). For literary and culturally dense texts, LLMs show potential in capturing imagery, metaphor, and cultural meaning, yet they also suffer from interpretive biases and limitations in stylistic reproduction (Zhang *et al.*, 2025). This study develops a role-conditioned and cross-validated LLM reader model, offering a more interpretable and cognitively aligned evaluation pathway than existing prompt-based assessment approaches.

## 3 Methods

This study uses prompts as a cognitive scaffold. The metaphoric and metonymic cues in TCM texts were revealed as a supplement, encoded into prompts for a large language model (LLM) to support the model's comprehension of diagnostic-therapeutic logic. Adjusting the model's reasoning patterns step by step, the LLM was guided to improve translations. Subsequently, two other LLMs were instructed through prompts to act as target-language readers and perform both quantitative and qualitative evaluations of the translations.

### *3.1 Selection of sample cases*

All four representative cases selected for the study were drawn from the canonical TCM work Huangdi Neijing. These passages are central to foundational TCM theory, complicated in interpretation, and notably difficult for English readers when rendered by literal translation. Each case is rich in metaphor and metonymy that are originated from the human–nature relationship, and form the basis of physiology, pathogenesis, and therapeutic principles in the narrative. However, the metaphoric projections and referential relations are not explicitly presented in the contexts and thus require supplementation by the translator. Through prompts these cues were supplemented to guide the model to draw the cognitive map and polish the translation. The cases are labeled Case 1 through Case 4. Case 1 focuses on conveying the metaphorical content embedded in terminology; Cases 2 and 3 address failures to reflect the projection and reference in the source texts; Case 4 additionally involves discussion of therapeutic principles and methods, requiring both metaphor–metonymy clarification and explicit explanation of treatment rationale.

### *3.2 Choice of LLMs*

As LLMs differ in style, preferences, and translation capability, a single LLM was adopted as the primary translation engine to ensure consistency, while employing multiple LLMs in the evaluation stage to assess cross-model agreement.

Deepseek V3.1 was used for cognitive prompt tuning. Prior benchmarks indicate Deepseek V3 series performs well on metaphor translation and produces consistent renderings (Li & Brom, 2025); its growing ecosystem (*e.g.*, DeepSeek-OCR) also suggests potential for larger-scale digitized medical texts.

For evaluation, ChatGPT-5 Pro and Gemini 2.5 Pro are used because of their widespread adoption in Anglophone contexts and their strong alignment with English readers' linguistic and reasoning habits, improving the authenticity of simulated reader judgments.

### *3.3 Prompt engineering*

Baseline translation and diagnostic test. Input the case in source language and prompt the model to generate a baseline translation and reasoning. Analyze the model's answer to determine whether: (a) the baseline sufficiently conveys the medical rationale; (b) shortcomings stem from lack of theoretical knowledge, inadequate metaphor/metonymy recognition and analysis, or purely linguistic limitations.

Injecting theoretical knowledge. If the model lacks domain background, supplement prompts with excerpts from primary texts and classical commentaries to provide needed theory.

Metaphor and metonymy identification. Augment prompts with explicit descriptions of source-domain imagery and metonymic referents, and explain how they build the diagnostic and therapeutic logic.

Polish the translation. (1) Instruct the model to preserve the source text's structural ordering while improving readability and concision; (2) encourage rendering imagistic meanings—*e.g.*, the metaphorical senses of qi movement—into dynamic verbal expressions in the translation.

### *3.4 Comparative evaluation of translations*

With tailored prompts, three types of English-speaking, real-world readers with distinct education and clinical backgrounds were simulated by LLMs:

Role1 (R1): A Western medical doctor interested in whole-system medicine who has attended an advanced international TCM clinical training program in China;

Role2 (R2): A U.S. licensed practitioner of Chinese medicine trained under NCCAOM standards;

Role3 (R3): A U.K. NHS physician who completed an MSc in TCM at a London college. (Full prompt templates are provided in the Appendix.)

For each case, the three simulated readers (run in both ChatGPT-5 Pro and Gemini 2.5 Pro) “evaluated” the four translations: the LLM baseline, the prompt-adjusted LLM translation (final), and two canonical human translations (Unschuld’s and Li Zhaoguo’s). The assignment of cases was single-blind: the four translations in each case were marked by randomized numbers to prevent model bias toward particular translators. The numbers were generated by Python using the Fisher–Yates algorithm as shown in Table 1:

Table 1. The randomized numbering of the translations in the four cases.

| **Case** | **Sequence*** | | | |
|---|---|---|---|---|
| Case 1 | 2 | 4 | 1 | 3 |
| Case 2 | 4 | 3 | 2 | 1 |
| Case 3 | 1 | 2 | 3 | 4 |
| Case 4 | 3** | 1 | 4 | 2 |

Code key: 1 = LLM baseline; 2 = LLM final; 3 = Unschuld; 4 = Li Zhaoguo.

(Note: Case 4 was extracted from *Huangdi Neijing Su Wen: Chapter 72*（《素问·刺法论》) and lacks a corresponding Unschuld version, so Li Zhaoguo’s translation substituted.)

A rating-discourse evaluation task was arranged for each simulated reader through prompt. For each case, they should rate the four translations on five cognitive dimensions—Clarity, Cognitive Load, Confidence in Understanding, Preference, and Clinical Transferability—using a 5-point Likert scale (with Cognitive Load scored positively from difficult to easy). After rating, the reader should: (1) paraphrase the medical rationale conveyed by each version; (2) identify unclear or confusing segments; and (c) explain how each translation informs their clinical thinking based on work experience and medical education.

### *3.5 Quantitative and qualitative analysis*

Quantitative analysis: as the ratings are ordinal (5-point scale) with a small sample, nonparametric tests were used. Spearman's $\rho$ and Kendall's $W$ were computed to assess inter-model and inter-role agreement. Friedman tests examine differences across translations; pairwise Wilcoxon signed-rank tests with Bonferroni correction are used for post-hoc comparisons to control for multiple testing. Significance thresholds were set at $p < 0.05$ (significant) and $p < 0.01$ (highly significant).

Qualitative analysis: guided by the quantitative outcomes, Interpretative Phenomenological Analysis (IPA) was applied following Smith *et al.*'s seven-step procedure (Smith *et al.*, 2022). Each simulated reader role formed a case analysis: the researcher immersed in the case-level experience, integrated researcher reflexivity, and synthesized emergent themes across cases to distill actionable translation strategies.

## 4 Results

We first examined the scoring consistency across the two LLMs and the three simulated reader roles. The results show high cross-model consistency (Spearman $\rho = 0.68$, $p < 0.01$; Kendall's $W = 0.79$, $p < 0.001$) and high cross-role consistency (GPT: $W = 0.73$, $\chi^2(15) = 32.85$, $p < 0.01$; Gemini: $W = 0.78$, $\chi^2(15) = 35.10$, $p < 0.01$). These patterns provide the methodological premise for subsequent analyses, ensuring the robustness and reproducibility of the findings.

### *4.1 Scores of the four translations across dimensions and roles*

Figure 2 presents the Likert-scale evaluations of the four translations (A = LLM baseline; B = LLM final; C = Unschuld; D = Li Zhaoguo) across five dimensions. The radar chart shows the general scoring tendencies, with the prompt-adjusted LLM translation outperforming all other versions on all dimensions. A Friedman test ($\chi^2 = 217.56$, $p < 0.001$) and Wilcoxon tests (Bonferroni-corrected) confirm significant differences among translation versions, indicating that cognitively oriented prompting substantially enhances translation quality for complex source texts. The largest gain appears in Clarity (mean: 3.91–4.58), whereas Cognitive Load shows the smallest change (mean: 3.54–3.58). Human translations display a larger gap from LLM translations across all dimensions, while differences within LLM translations (A vs. B) and within human translations (C vs. D) are comparatively small, suggesting different translational strategies between human and machine agents. These trends establish the groundwork for the qualitative analysis.

Figure 2. Average scores of four translation versions across five cognitive dimensions.

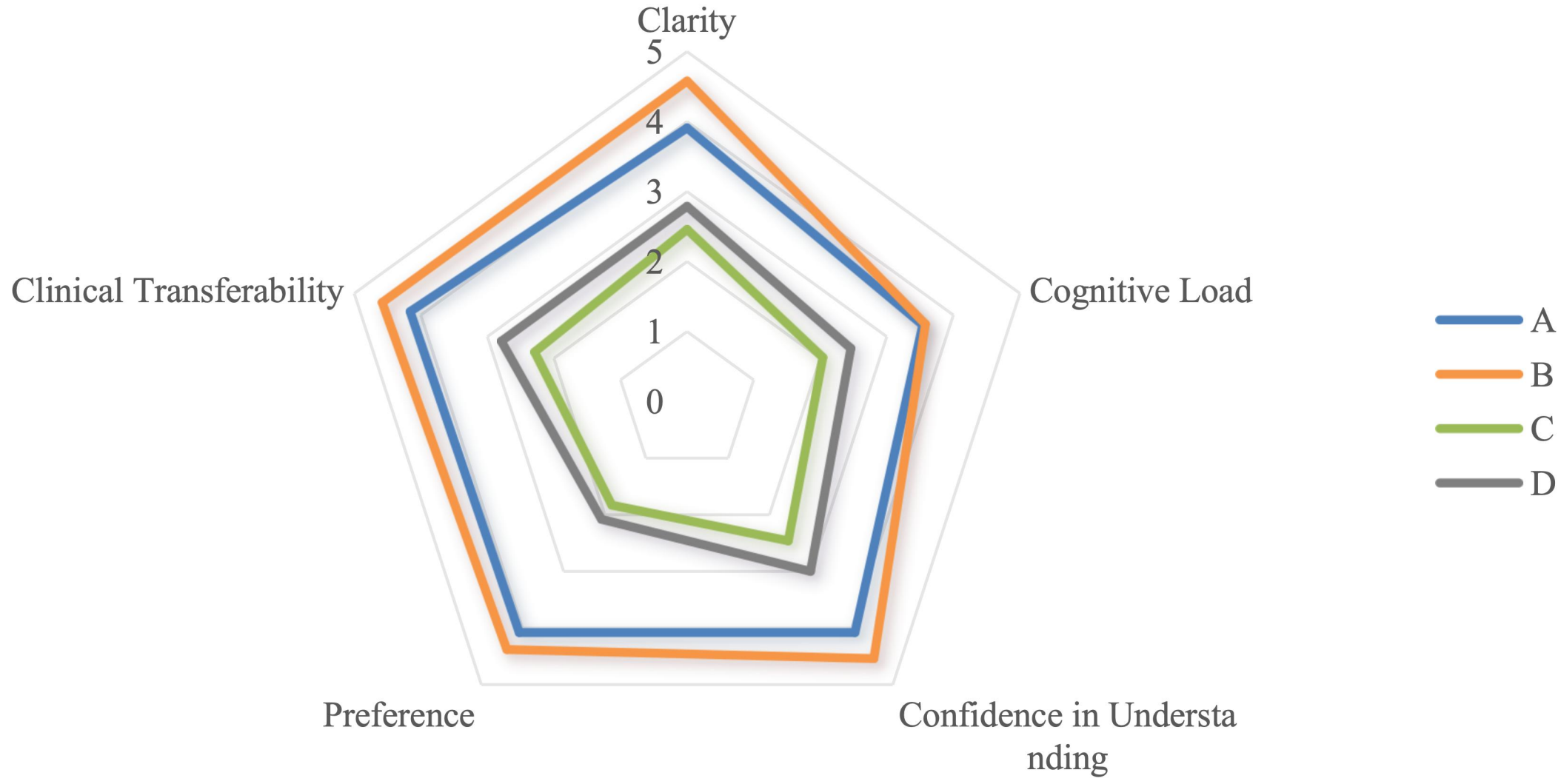

(Note: A= LLM baseline; B = LLM final; C = Unschuld; D = Li Zhaoguo)

Figure 3 shows the mean and range (max–min) of Likert scores given by the three simulated reader roles. Although the final LLM translation (B) receives the highest mean scores across all roles, its score range is wider than that of the baseline version (A), indicating less stability. All roles assign significantly lower scores to the human translations, but Role 2 (licensed TCM practitioner in the U.S.) exhibits a distinct pattern from the other two groups (Western clinicians with short-term TCM training; physicians with MSc-level TCM training), giving human translations a noticeably higher upper bound. These tendencies are revisited in the qualitative analysis.

Figure 3. Average scores of four translation versions by reader roles.

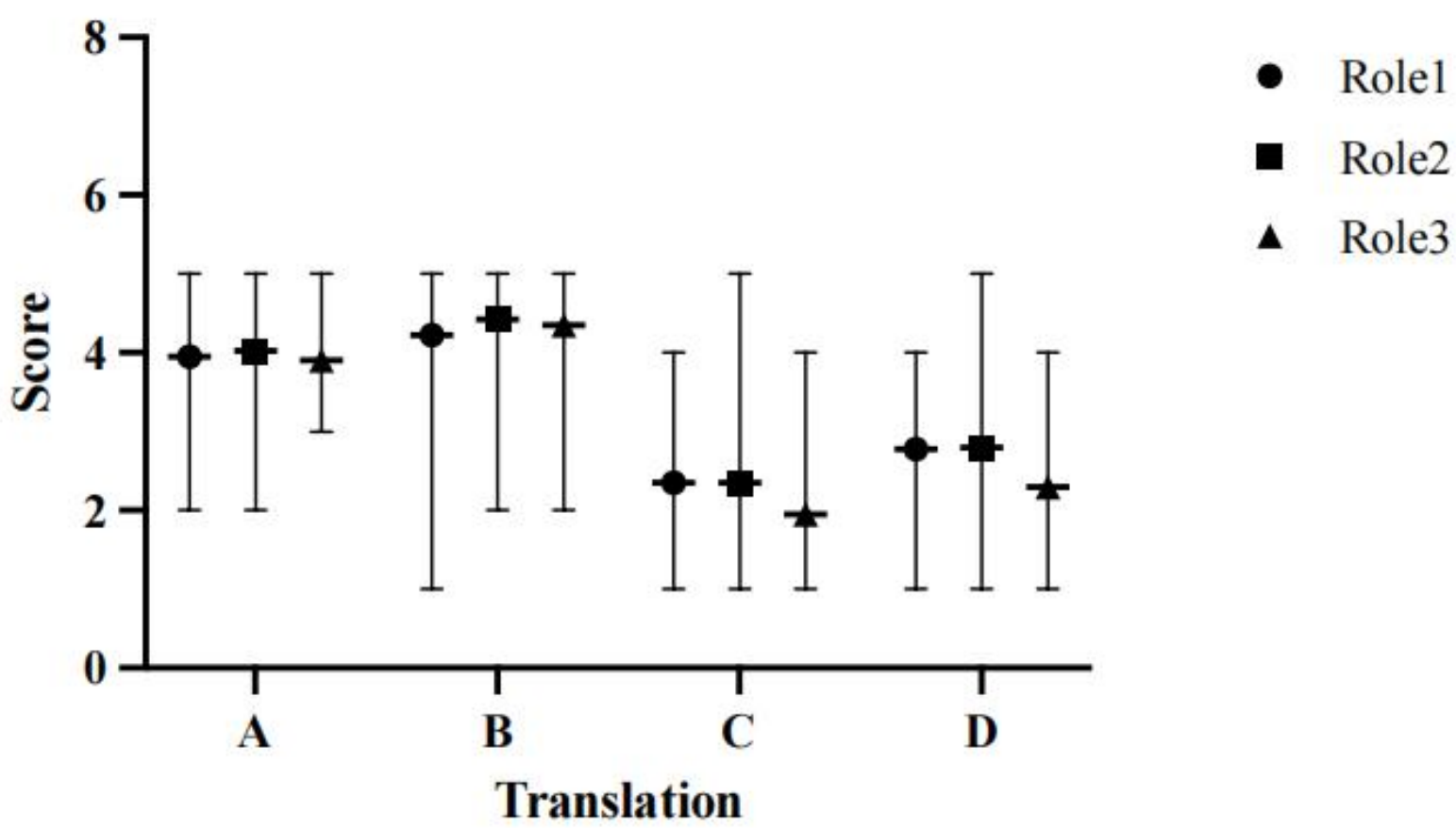

(Note: A= LLM baseline; B = LLM final; C = Unschuld; D = Li Zhaoguo)

### *4.2 Translation cases and prompt-guided adjustments*

This section illustrates how cognitively oriented prompting supports LLMs in identifying and handling complex metaphor and metonymy and in articulating the diagnostic logic embedded in the source text. Four thematic passages were selected: seasonal pathogenic wind (Case 1), the functionality and relations of the Five Organs (Case 2), the functional patterning of the meridians and the Six Qi (Case 3), and pathological effects arising from failed seasonal transitions (Case 4)—all core components of classical TCM theory. Each case presents (1) baseline translation problems, (2) prompt-adjustment strategies, (3) part of the final translation, and (4) corresponding Likert scores.

*Case 1 "Xuxie"（虚邪）: The contra-seasonal pathogenic wind defined by astronomical observation*

Original context: Wind blowing from a direction opposite to the one that normally dominates a given season harms the human body; sages avoid such wind.

Translation focus: Identifying the metaphorical domain: the Big Dipper sets the correct orientation associated with Taiyi, the cosmological principle that aligns with the season. Xuxie (void wind) refers to wind lacking the orientation of Taiyi, thereby functioning contra-seasonally and becoming pathogenic.

Problem in baseline translation: “Xu” was rendered as deficient, obscuring its metaphorical origin and likely to be confused with bodily deficiency.

Prompt strategy: (1) Provide passages from Huangdi Neijing as background knowledge to reconstruct the metaphorical mapping and have the LLM explicate the context. (2) Supply the metaphorical meaning of Xuxie and request a clinical-oriented rendering (*e.g.* contra-seasonal wind).

The comparison of the baseline and adjusted translations, along with their average evaluation scores, is provided in Table 2.

Table 2. Baseline and prompt-adjusted translation versions of Case 1 with their average evaluation scores.

| **Version** | **Translation (excerpt)** | **Average score** |
|---|---|---|
| Baseline | When the wind blows from the opposite direction (contrary to the seasonal norm), it is called the "deficient wind," which harms the human body and brings destruction and damage. | 4.87 |
| Adjusted | When the wind blows from directions unattended by Taiyi (contrary to the seasonal norm), it is called the "seasonal-opposing wind," which harms the human body and brings destruction and damage. | 3.50 |

*Case 2 Functionality and relations of the Five Organs*

Original context: Seasonal climate metaphors map the qi-transformation functions of the Five Organs onto Spring, Summer, Late Summer, Autumn, and Winter. These metaphors explain the generating–controlling cycles and underpin prognostic reasoning.

Translation focus: Avoid literal rendering and foreground dynamic functional processes.

Problem in baseline translation: Literal translation; metaphorical and metonymic relations between seasons, climates, and organs are not conveyed.

Prompt strategy: (1) Instruct the LLM that literal expressions like “seasons overcome each other” are unintelligible to readers; ask the model to interpret the season–climate–organ metaphor/metonymy. (2) Prompt the model to replace static verbs with dynamic ones reflecting qi-transformation.

The comparison of the two translations is displayed in Table 3.

Table 3. Baseline and prompt-adjusted translation versions of Case 2 with their average evaluation scores.

| Version | Translation (excerpt) | Average score |
| --- | --- | --- |
| Baseline | Summer overcomes Autumn,<br>Autumn overcomes Spring.<br>This is what is meant by "the conquest cycles of the five elements in their seasons." | 4.03 |
| Adjusted | Summer (Fire-Qi) ascends and thereby fuses the hardness of Autumn (Metal-Qi).<br>Autumn (Metal-Qi) gathers and thereby shapes the exuberance of Spring (Wood-Qi).<br>This is the dynamic balance known as "the conquest cycles of the Five Phases through the seasons. | 4.87 |

*Case 3 Functional interactions between the meridians and the Six Qi*

Original context: The six climatic qi describe the basic quality of meridian functions, reflecting the “root” (本) of them; yin/yang characteristics of meridian groupings constitute the apparent states and are defined as “branch” (标); zhongqi (中气) mediates, coordinates, and transforms between root and branch, explaining the qi interaction between paired interior(yin)–exterior(yang) meridians.

Translation focus: Clarifying the metaphoric meanings of root, branch, and mediating qi.

Problem in baseline translation: Literal translation obscures the functional interplay.

Prompt strategy: (1) Guide the LLM to translate a clinically oriented parallel passage and explain the medical reasoning of root–branch–mediating qi. (2) Internalize this metaphorical reasoning and transfer it to the current passage.

The comparison of the two translations is displayed in Table 4.

Table 4. Baseline and prompt-adjusted translation versions of Case 3 with their average evaluation scores.

| **Version** | **Translation (excerpt)** | **Average score** |
|---|---|---|
| Baseline | Above the Shaoyang (channel/system), the fire qi governs it; and its interior correspondence is seen in the Jueyin. | 3.16 |
| Adjusted | The Shaoyang system is fundamentally characterized by fire qi, manifests its functional activity as Shaoyang, and operates through the pivotal mediation of the Jueyin system. | 4.67 |

*Case 4 Pathological effects of impaired seasonal transitions*

Original context: When the annual motion (the overall quality of annual climate) obstructs the descent and operation of the seasonal climate, qi stagnation arises, guiding acupuncture strategies.

Translation focus: Multiple metaphorical and metonymic cues, including therapeutic implications.

Problem in baseline translation: Literal rendering; misinterpretation of dixuan (地玄), which metaphorically denotes “Water” in the Five Phases.

Prompt strategy: (1) Guide the model to identify the colour associated with xuan, its Five-Phase correspondence, and interpret the medical reasoning from the perspective of five motion and six qi (wuyun liuqi, 五运六气) theory. (2) Supply the metonymy mapping: Fire → seasonal function; dixuan → Water-dominated annual motion.

The comparison of the two translations is displayed in Table 5.

Excluding Case 1 (which will be discussed in the next chapter), all adjusted translations show substantial score increases, indicating that identifying and explicating metaphor and metonymy enhances translation quality and that prompt-based cognitive scaffolding is an effective strategy.

Table 5. Baseline and prompt-adjusted translation versions of Case 4 with their average evaluation scores.

| **Version** | **Translation (excerpt)** | **Average score** |
|---|---|---|
| Baseline | When fire qi desires to descend but is obstructed and suppressed by the earthly mysterious energy, it fails to enter despite its downward movement. | 3.83 |
| Adjusted | When the Fire Qi desires to descend and function, it is obstructed and suppressed by the Water-dominated Annual Motion (Zhongyun), represented by "Di Xuan". | 4.37 |

### *4.3 Thematic analysis of reader feedback*

To better understand target-reader needs and identify translation strategies suitable for complex TCM theory, this study collected structured interview data from the three simulated reader groups. An interpretative phenomenological analysis (IPA) yielded three overarching themes, which further validate the prompting framework and allow comparison of translational strategies between human and LLM translations.

*Theme 1: Features of human vs. LLM translations*

Readers generally found LLM translations more straightforward, whereas human translations imposed higher cognitive load. Literal translation was the key differentiator: human translators frequently rendered the text literally, producing concise and structurally faithful sentences, but readers reported that such versions conveyed definitional knowledge rather than clinical reasoning, making application difficult.

*Theme 2: Development of translation strategies*

During prompt adjustment, supplementing historical and textual context helped reconstruct metaphorical imagery and metonymic referents and integrate them into diagnostic logic. Readers reported improved confidence in understanding and applying the concepts. Furthermore, maintaining textual fluency and inserting metaphor/metonymy through natural syntactic choices (*e.g.*, dynamic verbs) reduced cognitive load and made the translations more vivid. However, providing extensive metaphorical background knowledge introduced cognitive burden and did not aid comprehension, suggesting limited usefulness of this strategy.

*Theme 3: Reader-specific cognitive features and evaluation orientations*

The three roles differ in educational background and interpretive orientation. These differences across roles are summarized in Table 6.

Cognitive load is the major point of divergence in their evaluations. Role 2, with only TCM-based training, perceives dense terminology not as unfamiliar concepts but as a rigorous anchor for comprehension, producing evaluations opposite to those of the other two roles.

Table 6. Reader roles and evaluation orientations.

| Role | Clinical context | TCM education | Evaluation focus |
|---|---|---|---|
| Role 1 | Western MD, integrative medicine | Short-term TCM training | Clinical applicability; conceptual linkage |
| Role 2 | Licensed TCM | Systematic TCM | Terminological rigour; |

| | practitioner (acupuncture) | theoretical training | alignment with classical theory |
| --- | --- | --- | --- |
| Role 3 | Western MD, interdisciplinary work | TCM MSc programme | Interdisciplinary integration |

Overall, the prompting strategy successfully accommodates the cognitive needs of different reader groups, approximates TCM-specific modes of reasoning, and achieves improved translation quality, reduced cognitive load, and enhanced explanatory and applicative value. It thus constitutes a viable and effective translation practice.

## 5. Discussion

### *5.1 Overview of major findings*

The simulated-reader evaluations show that cognitively informed prompt adjustments—particularly those that supplement metaphorical and metonymic cues—help reconstruct the text's original mode of reasoning and its imagistic explanation of medical principles in the Huangdi Neijing. By aligning the translation process with the narrative logic of "imagistic thinking" (象思维), the adjusted translations demonstrate consistent improvements across five cognitive–pragmatic dimensions—Clarity, Cognitive Load, Confidence in Understanding, Preference, and Clinical Transferability. Readers reported higher levels of comprehension, willingness to apply, and confidence in applying the translated content; their paraphrases of the medical rationale became more accurate, more coherent with classical TCM reasoning, and more conducive to integrating TCM insights into Western clinical thought.

Among the adjustment strategies, two were particularly well received: (1) mapping metaphorical meanings into verbal expressions to portray physiological and pathological dynamics vividly; and (2) refraining from directly adding the entire metaphorical or metonymic schema, instead guiding the model to analyse their explanatory function and transfer this function into diagnostic–therapeutic logic. These strategies allow the translation to balance fidelity to the source texts with clarity and conceptual concreteness in explicating medical principles.

### *5.2 Cognitive mechanisms underlying prompt engineering*

This study treats prompts as a cognitive scaffold that guides the LLM to reason about metaphor and metonymy through the Neijing's explanatory mode—its cosmological view of the human–nature relationship—and to transfer this reasoning appropriately into translation. Concretely, the prompts first examine the LLM's baseline and its inferential process, prompting it to explain the cognitive basis of its initial translation. This reveals the model's grasp of imagistic sources, metonymic relations, and textual features. Based on these observations, the researcher diagnoses whether deficiencies in the LLM's output stem from background-knowledge gaps, incomplete theoretical reasoning, or purely linguistic issues, and then revises the prompts accordingly.

For example, in Case 1 the LLM recognised the metaphoric origin of "xu xie" (虚邪) but failed to interpret and convey its theoretical meaning. The problem therefore lay in conceptual transfer rather than metaphor recognition, which informed subsequent prompt adjustments.

Three major cognitive pathways of prompt optimization emerged:

(1) Supplementing knowledge.

Prompts add contextual knowledge from other chapters of Huangdi Neijing and literature of the same era that describes the astronomical and agricultural practice, allowing the LLM to reconstruct the embodied cognition and historical worldview that shaped metaphor and metonymy. This enables identification of metaphoric projections and metonymic referents and supports an understanding of how these rhetorical devices ground physiological and pathological explanations. For example, prompts in Case4 encouraged the LLM to interpret the passage through the lens of five motion and six qi (wuyun liuqi, 五运六气) theory, then guided it heuristically to infer the metonymic relationship between xuan (玄), the colour black, and annual water qi motion, supported by excerpts from Huangdi Neijing such as “司气为玄化” and “司气者主岁同”.

(2) Transferring explanation.

Since the ultimate function of metaphor and metonymy is explanatory rather than decorative, prompts directed the LLM to centre medical rationale in the translation and retain the source’s concise style rather than explicate every imagistic link. In Case 4, the prompt “preserve the source structure as much as possible” guided the model to translate “地玄” directly as the annual water qi motion, avoiding excessive explication that would burden the reader. In Case 3, although the LLM correctly analysed the imagery behind root-branch-mediate qi (标本中气), it failed to represent their functional linkage; the prompt then instructed the model to focus on clinical significance rather than literal rendering.

(3) Constructing discourse.

This involves refining expression to make the medical logic vivid, coherent, and conceptually manageable, without introducing Western biomedical terminology or excessive pinyin (phonetic symbols of Chinese characters). In Case 3, the initial adjusted version abandoned the original structure and produced a purely expository explanation. A prompt then required the model to follow the source text structure while integrating the earlier theoretical analysis, resulting in a translation that extends the source text discursively without becoming a total commentary. In Case 2, literal rendering reduced the passage to a mechanical model; prompts encouraged the model to express the Five Phases' interactions through dynamic verbs, capturing their generative and regulatory functions more effectively.

### *5.3 Interpretation of emergent themes from reader feedback*

*Translation strategies requiring refinement*

Readers consistently indicated that direct, literal translation should be avoided in TCM classics. Literal renderings obscure the intended medical logic and may mislead practitioners, reducing confidence in clinical applicability. For example, in Case 3, comments included:

R1: "'Ceasing yin' (厥阴) is awkward: does yin stop/transform at the midpoint, or does yin become quiescent? hard to map onto clinical signs."

R2: "Not interpretable for clinical use. Over-literal, impractical. Stiff, confusing syntax."

In contrast, readers preferred plain, familiar English phrasing—even when this meant the baseline translation outperformed the adjusted version in Case 1. Pinyin, the phonetic symbols of Chinese characters, should be used only sparingly, ideally with the Chinese text and a brief definition in parentheses to avoid homophone ambiguity. For example, R2 misinterpreted "sheng (胜)" as "sheng (生)" in Li Zhaoguo's translation of Case 2.

Background knowledge supporting metaphor and metonymy should be added flexibly. In Case 1, the final translation scored lower because cosmological notions such as Taiyi were perceived as culturally distant and cognitively burdensome.

R1 commented: “I must ignore the astronomical detail and focus solely on the ‘correct vs. opposite’ dichotomy.”

Although the baseline version mishandled terminology, its medical meaning was clear enough for readers to restate accurately and apply clinically. This suggests that terminology itself may not serve as the primary cognitive anchor; rather, it is the surrounding conceptual context that plays this role. Translators should therefore prioritise explanatory coherence over exhaustive background supplementation; such supplementation that describes the metaphoric origin could be placed in footnotes rather than integrated into the main text.

*Effective translation strategies*

Embedding metaphorical meaning into verbs greatly helps readers envision physiological and pathological dynamics and grasp inter-category relationships. In Case 2, this strategy clarified Five-Phase interactions, prompting R3 to observe:“Although the longest, it is the easiest to process… the translator explicitly links the Phase, the Season, and the mechanism.”

Where metaphorical and metonymic cues were fully restored (Cases 3 and 4), all readers reported increased confidence in clinical applicability. R2 noted in Case 3:“I’d trust this version only—it guides diagnostic thinking…the Shaoyang (Gallbladder/Sanjiao) expresses Fire, its internal coordination with Jueyin (Liver) governs the pivot between exterior and interior—key in alternating heat–cold or Shaoyang disorders.”

R1 evaluated Case 4 similarly, calling human translations “context-poor” and the baseline AI translation “vague,” whereas the adjusted version was “logically complete” despite some cognitive load from five motions and six qi (wuyun liuqi,五运六气) terminology. For R1, the supplemented metonymic relations—such as the one between xuan and annual water qi motion—were indispensable for understanding the medical logic.

Cognitive needs of different reader profiles

Reader roles exhibited distinct cognitive orientations:

R1 (Western clinician with TCM training) emphasised clinical applicability and connections to Western concepts such as endocrine regulation, homeostasis, and circadian rhythms. For Case 2, he likened Five-Phase regulation to counter-regulatory hormone dynamics:

> "Regulatory Mechanisms: This strongly aligns with my understanding of metabolic regulation and counter-regulatory hormones (*e.g.*, insulin vs. glucagon). It suggests a system where excesses in one area are managed by the active mechanism of another. It's an analogy for physiological checks and balances."

R2 (TCM practitioner) prioritised terminological precision and doctrinal coherence. Standardised TCM terminology reduced cognitive load and activated familiar theoretical networks, such as his spontaneous association of Case 1 with the liuyin (六淫) framework. Standardized TCM terminology prompts associations with R2's prior training in systematic TCM theory, which helps to lower cognitive load：

"Uses the most standard and recognizable TCM terminology ("substantial wind," "deficient wind") making the meaning instantly accessible and clinically relevant. Low cognitive load."

R3 (NHS physician with MSc in TCM) demonstrated an interdisciplinary, conceptual approach. In interpreting Case 4, he drew parallels between TCM "fire constraint" and modern models of dysregulated metabolic flow, noting the explanatory power of metaphor across medical paradigms：

> "Within modern pathological frameworks — such as hypertension, anxiety, and metabolic syndrome — the traditional notion of 'fire that wishes to descend but is constrained' can metaphorically represent the coexistence of blocked energy metabolism and emotional stagnation. Thus, it enriches the metaphorical language used in clinical interpretation and also suggests that regulation should not be seen merely as inhibition or activation, but rather as the restoration of flow — of movement, circulation, and descent."

Notably, all three readers displayed reflective reasoning that moved beyond assimilating TCM into Western thought; they actively explored bidirectional conceptual integration. For instance, R1 commented that TCM theory encourages attention to the "quality of exposure (It's not just the wind; it's the wind that is 'deficient' or 'contra-seasonal')", a nuance largely absent in Western environmental etiology.

Such responses further support the value of cognitively informed prompt adjustment: it not only clarifies TCM theory but also cultivates readers' imagistic thinking, enabling deeper cross-system dialogue and more effective clinical practice.

## 6. Conclusion

This study employed prompt-based guidance to enable LLMs to identify and transfer metaphorical and metonymic cues in classical TCM texts, while using LLM-simulated readers to perform a mixed-methods evaluation across five cognitive dimensions. The approach demonstrably enhanced translation quality and offers a novel pathway for both the translation and assessment of TCM classics.

From the perspective of methodological optimization, the iterative prompt refinement process guided the LLM in recognizing metaphorical and metonymic logic as well as their narrative functions, transferring these insights into explanatory translations. This process simulates the construction of meaning in TCM thought through imagistic schemata, reflecting how physiological and pathological mechanisms are interpreted in traditional texts. Unlike standard terminological or knowledge-based translation approaches, this method seeks cognitively aligned source-target mappings, extending the capabilities of vector-based representations to facilitate readers' engagement with the distinct features of TCM imagistic thinking. Such alignment can support clinical practice and substantially improve the translatability of TCM texts. Although the study is limited to a small number of cases, these segments are highly cognitively complex and theoretically representative, suggesting that the method could be scaled to generate fine-tuning datasets, enhance LLM cognitive abilities, and support large-scale translation of TCM texts. Given that English remains the dominant language for TCM translation, evaluations in this study focused primarily on English (Wang & Chen, 2023). The source-text-centered optimization method transcends the linguistic characteristics of the target language, offering a transferable cognitive-tuning paradigm applicable to other languages, demonstrating its forward-looking potential.

Regarding evaluation, LLM-simulated readers proved to be an efficient and stable tool. Trained on extensive natural language corpora, LLM-generated assessments are broadly generalizable, reducing the time and cost of feedback collection, expanding the scope and dimensions of comparative evaluation, and enabling translators to rapidly determine basic strategies based on readers' pragmatic needs before initiating TCM translations. This method also supports large-scale, longitudinal comparisons of translations. Notably, the LLM stimulates an *idealized* reader role (*e.g.*, a hypothetical English clinician who has *fully mastered* TCM theory), and its feedback inevitably differs from that of human readers. It cannot entirely replace the real-world quantitative and qualitative research. Future work should continue to refine and validate LLM-generated translations through real-world empirical review and assessment.

Acknowledgements

The author thanks the anonymous reviewers and the editorial team for their valuable comments. This research received no specific grant from any funding agency.

## Appendix:

## Prompt for evaluation:

You are now playing the role of

[A) A Western-trained physician interested in integrative medicine who has attended an International Advanced Training Program on Clinical Practice and Research Progress in Traditional Chinese Medicine in China;
B) A licensed TCM practitioner in the United States who has received NCCAOM-accredited TCM training;
C) A Western-trained physician working in the UK NHS system who has completed a Master's program in Chinese Medicine at the London Chinese Medicine College.]
Your native language is English, and you should fully immerse yourself in this role.

After reading the following four English translations of passages from the *Huangdi Neijing (黄帝内经)*, complete the tasks below and demonstrate your actual process of understanding:

***1. Degree of understanding and points of confusion:***

After reading the four translations, which translation gave you a clearer understanding of the following concepts:

• the nature of term *虚邪* (contra-seasonal pathogenic qi)/

• the functional relationships among the five organs across the four seasons/

• the nature of ***标本中气****(root/ branch/ mediating qi of the meridians)*

• the patterns of interaction between the qi of Heaven and Earth and the related mechanisms of disease?

Rate each of the four translations (1 2 3 4 5: unclear – somewhat unclear – average – fairly clear – very clear).

For each translation, circle all the words or expressions that you find confusing, unnatural, or not immediately understandable. Explain the specific reasons for each point of confusion.

***2. Concept restatement and meaning construction***

Using your own words, restate the main theoretical ideas presented in each of the four translations.

During this restatement, in what areas did you find yourself "filling in the gaps" or making guesses?

Which medical or health-related concepts that you already know did these translations make you think of?

***3. Cognitive load***

Rate the difficulty of understanding each of the four translations, noting that the rating is relative **within the domain of TCM theory**, not in comparison to casual or popular texts.
(1 2 3 4 5: difficult – somewhat difficult – average – relatively easy – very easy)

Where do the main sources of difficulty among the four translations come from?

***4. Confidence in understanding***

Overall, how confident are you in your understanding?
(1 2 3 4 5: no confidence – low confidence – average – fairly confident – very confident)

If you had to make clinical decisions—such as explaining to a patient or forming a treatment plan—based on this understanding, would you feel comfortable?

***5. Translation preference***

Personally, which translation would you prefer to use when explaining the concepts to your patients or colleagues?
Rate each translation (1 2 3 4 5: would never choose – unlikely to choose – neutral – somewhat likely – very likely).

Why do you have this preference?

***6. Transferability of theory to clinical practice***

To what extent, and in what way, would you apply the theories in the translations to real diagnostic or therapeutic situations?
Rate each translation (1 2 3 4 5: would not apply at all – unlikely to apply – neutral – likely to apply – definitely would apply), and explain how you would apply it.

Did this text change your understanding of health or disease? If so, how?